\documentclass[runningheads]{llncs}
\usepackage{graphicx}

\usepackage{xcolor}
\PassOptionsToPackage{hyphens,spaces}{url}
\usepackage[colorlinks,urlcolor=blue]{hyperref}
\usepackage{todonotes}
\usepackage{xspace}
\usepackage{bm}
\usepackage[utf8]{inputenc}
\usepackage{amssymb}
\usepackage{amsfonts}
\usepackage{amsmath}

\usepackage{graphicx}
\usepackage{booktabs}
\usepackage{footnoterange}
\usepackage{multirow}

\usepackage{siunitx}

\newcommand{\C}{\mathcal{C}}
\newcommand{\D}{\mathcal{D}}
\newcommand{\N}{\mathcal{N}}
\newcommand{\PosC}{\C^{+}}
\newcommand{\NegC}{\C^{-}}

\newcommand*{\M}{\mathcal{M}}

\renewcommand*{\S}{\mathcal{S}}
\renewcommand*{\P}{\mathcal{P}}
\newcommand*{\T}{\mathcal{T}}
\newcommand*{\para}[1]{\noindent\textbf{#1}}

\begin{document}
\title{ENIGMA Anonymous:\\ Symbol-Independent Inference Guiding
  Machine\\(system description)
  \thanks{Supported by the
    ERC Consolidator grant AI4REASON
    no.~649043 %
    (JJ, BP, MS, and JU), the Czech project AI\&Reasoning
    CZ.02.1.01/0.0/0.0/15\_003/0000466 and the European Regional
    Development Fund (KC and JU), the ERC Project \emph{SMART}
    Starting Grant no.~714034 (MO), grant 2018/29/N/ST6/02903 of
    National Science Center, Poland (BP), and the Czech Science
    Foundation project 20-06390Y (MS).}}
\titlerunning{ENIGMA Anonymous}
\author{Jan Jakub\r{u}v\inst{1}
\and Karel Chvalovsk\'y\inst{1}
\and  Miroslav Ol\v{s}\'ak\inst{2}
\and  Bartosz Piotrowski\inst{1,3}
\and Martin Suda\inst{1}
\and  Josef Urban\inst{1}}
\authorrunning{Jan Jakub\r{u}v et al.}
\institute{
  Czech Technical University in Prague, Czechia \\
\and
University of Innsbruck
\and
University of Warsaw
}

\maketitle              %
\begin{abstract}
We describe an implementation of gradient boosting and neural
guidance of saturation-style automated theorem provers that does
not depend on consistent symbol names across problems. For the gradient-boosting
guidance, we manually create abstracted features by considering
arity-based encodings of formulas. For the neural guidance, we
use symbol-independent graph neural networks (GNNs) and their embedding
of the terms and clauses. The two methods are efficiently implemented in the
E prover and its ENIGMA learning-guided framework.

To provide competitive real-time performance of the GNNs, we have developed
a new context-based approach to evaluation of generated clauses in E.
Clauses are evaluated jointly in larger batches and with respect to a large number of already
selected clauses (context) by the GNN that estimates their collectively most useful subset in several rounds of message passing.
This means that approximative inference rounds done by the GNN are efficiently interleaved with precise symbolic inference rounds done inside E.
The methods are evaluated on the MPTP large-theory benchmark and
shown to achieve comparable real-time performance to state-of-the-art symbol-based methods.
The methods also show high complementarity, solving a large number of hard Mizar problems.
\end{abstract}

\section{Introduction: Symbol Independent Inference Guidance }
\label{sec:intro}

In this work, we develop two \emph{symbol-independent} (anonymous) inference
guiding methods for saturation-style automated theorem provers (ATPs)
such as E~\cite{Sch02-AICOMM} and Vampire~\cite{Vampire}. Both methods
are based on learning clause classifiers from previous proofs within
the ENIGMA framework~\cite{JakubuvU17a,JakubuvU18,ChvalovskyJ0U19}
implemented in E. By \emph{symbol-independence} we mean that no
information about the symbol names is used by the learned guidance. In
particular, if all symbols in a particular ATP problem are
consistently renamed to new symbols, the learned guidance will result
in the same proof search and the same proof modulo the renaming.

Symbol-independent guidance is an important challenge for learning-guided ATP, addressed
already in Schulz's early work on learning guidance in
E~\cite{DBLP:books/daglib/0002958}. With ATPs being increasingly used
and trained on large ITP
libraries~\cite{hammers4qed,BlanchetteGKKU16,holyhammer,KaliszykU13b,DBLP:journals/jar/CzajkaK18,DBLP:conf/cpp/GauthierK15},
it is more and more rewarding to develop methods that learn to reason
without relying on the particular terminology adopted in a single
project. Initial experiments in this direction using concept
alignment~\cite{DBLP:journals/jsc/GauthierK19} methods have already
shown performance improvements by transferring knowledge between the
HOL libraries~\cite{DBLP:conf/lpar/GauthierK15}. Structural analogies
(or even terminology duplications) are however common already in a
single large ITP library~\cite{ckju-mcs-hh} and their automated
detection can lead to new proof ideas and a number of other interesting
applications~\cite{GauthierKU16}.

This system description first briefly introduces saturation-based ATP
with learned guidance (Section~\ref{sec:atp+ml}).  Then we discuss
symbol-independent learning and guidance using abstract features and gradient boosting trees 
(Section~\ref{sec:gbdt}) and graph neural networks
(Section~\ref{sec:gnn}). The implementation details are explained in
Section~\ref{sec:perf} and the methods are evaluated on the MPTP
benchmark in Section~\ref{sec:exp}.

\section{Saturation Proving  Guided by Machine Learning}
\label{sec:atp+ml}

\para{Saturation-based Automated Theorem Provers} 
(ATPs) such as E and Vampire are used to prove goals $G$ using a set of axioms $A$.
They clausify the formulas $A\cup\{\lnot G\}$ and 
try to deduce contradiction using the \emph{given clause loop}~\cite{Overbeek:1974:NCA:321812.321814} as follows.
The ATP maintains two sets of processed ($P$) and unprocessed ($U$) clauses. 
At each loop iteration, a given clause $g$ from $U$ is selected, moved to $P$,
and $U$ is extended with new inferences from $g$ and $P$.
This process continues until the contradiction is found, $U$ becomes empty, or a
resource limit is reached. The search space %
grows quickly and selection of the
right given clauses is critical. %

\vspace{1mm}
\para{Learning Clause Selection} over a set of related problems is a general method how to guide the proof search.
Given a set of FOL problems $\P$ and initial ATP strategy $\S$, we can
evaluate $\S$ over $\P$ obtaining training samples $\T$.
For each successful proof search, training samples $\T$ contain the set of
clauses processed during the search.
\emph{Positive} clauses are those that were \emph{useful} for the proof search (they appeared in
the final proof), while the remaining clauses were \emph{useless}, forming the \emph{negative} examples.
Given the samples $\T$, we can \emph{train} a machine learning \emph{classifier} $\M$ which
predicts usefulness of clauses in future proof searches.
Some clause classifiers are described in detail in Sections~\ref{sec:gbdt}, \ref{sec:gnn}, and \ref{sec:perf}.

\vspace{1mm}
\para{ATP Guidance By a Trained Classifier:}
Once a clause classifier $\M$ is trained, we can use it inside an ATP.
An ATP strategy $\S$ is a collection of proof search parameters such as
term ordering, literal selection, and also given clause selection
mechanism.
In E, the given clause selection is defined by a collection of clause \emph{weight
functions} which alternate to select the given clauses.
Our ENIGMA framework uses two methods of plugging the trained classifier $\M$ into $\S$.
Either (1) we use $\M$ to select all given clauses (\emph{solo mode} denoted $\S\odot\M$), or (2) we combine
predictions of $\M$ with clause selection mechanism from $\S$ so that roughly
$50\%$ of the clauses is selected by $\M$ (\emph{cooperative mode} denoted
$\S\oplus\M$).
Proof search settings other than clause selection are inherited from $\S$ in
both the cases.
See~\cite{ChvalovskyJ0U19} for %
details. The phases of learning and ATP guidance can be iterated in a \emph{learning/evaluation loop}~\cite{US+08-long}, yielding growing sets of proofs $\T_i$ and stronger classifiers $\M_i$ trained over them. See~\cite{JakubuvU19} for such large experiment.

\section{Clause Classification by Decision Trees}
\label{sec:gbdt}

\para{Clause Features} are used by ENIGMA to represent clauses as sparse vectors for machine learners.
They are based mainly on
vertical/horizontal cuts of the clause syntax tree.
We use
simple \emph{feature hashing} to %
handle theories with large number
of symbols.
A clause $C$ is represented by the vector $\varphi_C$ whose $i$-th index
stores the value of a feature with hash index $i$.
Values of conflicting features (mapped to the same index) are summed.
Additionally, we embed \emph{conjecture features} into the clause representation
and we work with vector pairs $(\varphi_C,\varphi_G)$ of size $2*\mathit{base}$,
where $\varphi_G$ is the feature vector of the current goal (conjecture).
This allows us to provide goal-specific predictions.
See \cite{JakubuvU19} for more details.

\vspace{1mm}
\para{Gradient Boosting Decision Trees (GBDTs)} implemented by the
XGBoost library~\cite{Chen:2016:XST:2939672.2939785} currently provide the strongest
ENIGMA classifiers. Their speed is comparable to the previously used~\cite{JakubuvU18} weaker 
linear logistic classifier, implemented by the LIBLINEAR
library~\cite{Fan:2008:LLL:1390681.1442794}.
In this work, we newly employ the LightGBM~\cite{LightGBM} GBDT 
implementation.
A \emph{decision tree} is a binary tree whose nodes contain Boolean conditions
on values of different features.
Given a feature vector $\varphi_C$, the decision tree can be navigated from the
root to the unique tree leaf which contains the classification of clause
$C$.
GBDTs combine predictions from a collection of follow-up decision trees.
While inputs, outputs, and API of XGBoost and LightGBM are compatible, each
employ a different method of tree construction.
XGBoost constructs trees level-wise, while LightGBM leaf-wise.
This implies that XGBoost trees are well-balanced. %
On the other hand, LightGBM can produce much deeper trees and the tree depth
limit is indeed an important learning meta-parameter which must be additionally
set.

\vspace{1mm}
\para{New Symbol-Independent Features:}
We develop a feature anonymization method based on
symbol arities. 
Each function symbol name $s$ with arity $n$ %
is
substituted by a special name ``\texttt{f}$n$'', while a predicate symbol
name $q$ with arity $m$ is substituted by ``\texttt{p}$m$''.
Such features lose the ability to distinguish different symbol names, and
many features are merged together.
Vector representations of two clauses with renamed symbols are clearly equal.
Hence the underlying machine learning method will provide equal predictions for such clauses.
For more detailed discussion and comparison with related work see Appendix~\ref{app:anon}.

\vspace{1mm}
\para{New Statistics and Problem Features:}
To improve the ability to distinguish different anonymized clauses,
we add the following features.
\emph{Variable statistics} of clause $C$ containing 
(1) the number of variables in $C$ without repetitions, 
(2) the number of variables with repetitions, %
(3) the number of variables with exactly one occurrence,
(4) the number of variables with more than one occurrence, 
(5-10) the number of occurrences of the
most/least (and second/third most/least) occurring variable.
\emph{Symbol statistics} do the same for symbols instead of variables.
Recall that we embed conjecture features in clause vector pair
$(\varphi_C,\varphi_G)$.
As $G$ embeds information about the conjecture but not about the problem
axioms, we propose to additionally embed some statistics of the problem $P$
that $C$ and $G$ come from.
We use 22 problem features
that E prover already computes for each input
problem to choose a suitable strategy.
These are (1) number of goals, (2) number of axioms, (3) number of unit
goals, etc.
See E's manual for more details.
Hence we work with vector triples $(\varphi_C,\varphi_G,\varphi_P)$.

\section{Clause Classification by Graph Neural Network}
\label{sec:gnn}

Another clause classifier newly added to ENIGMA is based on graph
neural networks (GNNs). We use the symbol-independent network architecture developed
in~\cite{DBLP:journals/corr/abs-1911-12073} for premise
selection. As~\cite{DBLP:journals/corr/abs-1911-12073} contains all
the details, we only briefly explain the basic ideas behind this
architecture here.

\vspace{1mm}
\para{Hypergraph.}
Given a set of clauses $\mathcal C$ we create a directed hypergraph with three
kinds of nodes that correspond to clauses, function and predicate
symbols $\mathcal N$, and unique (sub)terms and literals $\mathcal U$
occurring in $\mathcal C$, respectively. There are two kinds of hyperedges
that describe the relations between nodes according to $\mathcal
C$. The first kind encodes literal occurrences in clauses
by connecting the corresponding nodes. The second hyperedge kind
encodes the relations between nodes from
$\mathcal N$ and $\mathcal U$. For example, for
$f(t_1,\dots,t_k)\in\mathcal U$ we loosely speaking %
connect the nodes $f\in\mathcal N$ and
$t_1,\dots,t_k\in\mathcal U$ with the node $f(t_1,\dots,t_k)$ and
similarly for literals, where their polarity is also taken into
account.

\vspace{1mm}
\para{Message-passing.}
The hypergraph describes the relation between various kinds of objects
occurring in $\mathcal C$. Every node in the hypergraph is initially assigned a
constant vector, called the \emph{embedding}, based only on its kind ($\mathcal C$, $\mathcal N$, or $\mathcal U$).
These node embeddings are updated in a fixed
number of message-passing rounds, based on the embeddings of each node's neighbors.
The underlying idea of such neural message-passing
methods\footnote{Graph convolutions are a generalization of
  the sliding window convolutions used for
  aggregating neighborhood information in neural networks used for
  image recognition.} is to make the node embeddings encode more and
more precisely the information about the connections (and thus various properties)
of the nodes.
For this to work, we have to learn initial
embeddings for our three kinds of nodes and the update
function.\footnote{We learn individual components, which correspond to
  different kinds of hyperedges, from which the update function is
  efficiently constructed.}

\vspace{1mm}
\para{Classification.}
After the message-passing phase, the final clause embeddings are available
in the corresponding clause nodes. The estimated probability of a
clause being a good given clause is then computed by a neural network
that takes the final embedding of this clause and also aggregated
final embeddings of all clauses obtained from the negated conjecture.

\section{Learning and Using the Classifiers, Implementation}
\label{sec:perf}
In order to use either GBDTs (Section~\ref{sec:gbdt}) or GNNs 
(Section~\ref{sec:gnn}), a prediction model must be learned. 
Learning starts with training samples $\T$, that is, a set of
pairs $(\PosC,\NegC)$ of positive and negative clauses.
For each training sample $T\in\T$, we additionally know the source problem $P$
and its conjecture $G$.
Hence we can consider one sample $T\in\T$ as a quadruple 
$(\PosC,\NegC,P,G)$ for convenience.

\vspace{1mm}
\para{GBDT.}
Given a training sample $T=(\PosC,\NegC,P,G)\in\T$, each clause
$C\in\PosC\cup\NegC$ is translated to the feature vector
$(\varphi_C,\varphi_G,\varphi_P)$.
Vectors where $C\in\PosC$ are labeled as positive, and otherwise as
negative.
All the labeled vectors are fed together to a GBDT trainer yielding model
$\D_\T$.

When predicting a generated clause, the feature vector is computed and $\D_\T$ is asked
for the prediction.
GBDT's binary predictions (positive/negative) are turned into E's clause weight
(positives have weight $1$ and negatives $10$).

\vspace{1mm}
\para{GNN.}
Given $T=(\PosC,\NegC,P,G)\in\T$ as above we construct a hypergraph for the
set of clauses $\PosC\cup\NegC\cup G$.
This hypergraph is translated to a tensor representation (vectors and
matrices), marking clause nodes as positive, negative, or
goal.
These tensors are fed as input to our GNN training, yielding a GNN
model $\N_\T$.
The training works in iterations, and $\N_\T$ contains one GNN per
iteration epoch.
Only one GNN from a selected epoch is used for predictions during the
evaluation.

In evaluation, it is more efficient to compute predictions for several clauses
at once.
This also improves prediction quality as the queried
data resembles more the training hypergraphs where multiple clauses 
are encoded at once as well.
During an ATP run on problem $P$ with the conjecture $G$, we postpone
evaluation of newly inferred clauses until we reach a certain amount of clauses
$\mathcal{Q}$ to \emph{query}.\footnote{We may evaluate less than $\mathcal{Q}$ if E runs out of unevaluated unprocessed clauses.}  
To resemble the training data even more, we add a fixed number
of the %
given clauses processed so far. %
We call these \emph{context} clauses ($\mathcal{X}$).
To evaluate $\mathcal{Q}$, we construct the hypergraph for $\mathcal{Q}\cup\mathcal{X}\cup G$,
and mark clauses from $G$ as goals.
Then model $\N_\T$ is asked for predictions on $\mathcal{Q}$ (predictions for
$\mathcal{X}$ are dropped).
The numeric predictions computed by $\N_\T$ are directly used as E's weights.

\vspace{1mm}
\para{Implementation \& Performance.}
We use GBDTs implemented by the XGBoost~\cite{Chen:2016:XST:2939672.2939785} and
LightGBM~\cite{LightGBM} libraries.
For GNN we use Tensorflow~\cite{tensorflow2015-whitepaper}.
All the libraries provide Python interfaces and C/C++ APIs.
We use the Python interfaces for training and the C APIs for the evaluation in E.
The Python interfaces for XGBoost and LightGBM include the C APIs, %
while for Tensorflow this must be manually compiled,
which is further complicated by poor documentation.

The libraries support training both on CPUs and on GPUs.
We train LightGBM on CPUs, and XGBoost and Tensorflow on GPUs.
However, we always evaluate on a single CPU as we aim at practical usability 
on standard hardware. 
This is non-trivial and it distinguishes this work from evaluations done with
large numbers of GPUs or TPUs and/or in prohibitively high real times.
The LightGBM training can be parallelized much better --  
with 60 CPUs it is much faster than XGBoost on 4 GPUs. 
Neither using GPUs for LightGBM nor many CPUs for XGBoost provided better
training times.
The GNN training is slower than GBDT training and
it is not easy 
to make Tensorflow evaluate reasonably
on a single CPU.
It has to be compiled with all %
CPU
optimizations and restricted to a single thread, using Tensorflow's poorly documented experimental C API.
\section{Experimental Evaluation}
\label{sec:exp}

\para{Setup.}
We experimentally evaluate\footnote{On a server with 36 hyperthreading Intel(R)
Xeon(R) Gold 6140 CPU @ 2.30GHz cores, 755 GB of memory, and 
4 NVIDIA GeForce GTX 1080 Ti GPUs.}
our GBDT and GNN guidance\footnote{
Available at \url{https://github.com/ai4reason/eprover-data/tree/master/IJCAR-20}}
on a large benchmark of $57880$
Mizar40~\cite{KaliszykU13b} problems%
\footnote{\url{http://grid01.ciirc.cvut.cz/~mptp/1147/MPTP2/problems_small_consist.tar.gz}}
exported by MPTP~\cite{Urban06}.
Hence this evaluation is compatible with our previous symbol-dependent
work~\cite{JakubuvU19}.
We evaluate GBDT and GNN separately.
We start with a good-performing E strategy $\S$
(see~\cite[Appendix~A]{ChvalovskyJ0U19}) which solves \num{14966} problems with a
\SI{10}{\second} limit per problem.
This gives us training data $\T_0=\mathsf{eval}(\S)$ (see Section~\ref{sec:perf}), 
and we start three iterations of the learning/evaluation loop (see Section~\ref{sec:atp+ml}).

For GBDT, we train several models (with hash base $2^{15}$) and conduct a small learning
meta-parameters \emph{grid search}.
For XGBoost, we try different tree depths ($d\in\{9,12,16\}$), and for LightGBM
various combinations of tree depths and leaves count 
($(d,l)\in\{10,20,30,40\}\times\{1200,1500,1800\}$).
We evaluate all these models in a cooperative mode with $\S$ on a 
random (but fixed) $10\%$
of all problems (Appendix \ref{app:data}).
The best performing model is %
evaluated on the whole benchmark in both cooperative ($\oplus$) and solo
($\odot$) runs.
These give us the next samples $\T_{i+1}$. %
We perform three iterations and obtain models $\D_0$, $\D_1$, and $\D_2$.

For GNN, we train a model with 100 epochs, obtaining $100$ different GNNs.
We evaluate GNNs from selected epochs ($e\in\{10,20,50,75,100\}$) and we try
different settings of \emph{query} ($q$) and \emph{context} ($c$) sizes (see
Section~\ref{sec:perf}).
In particular, $q$ ranges over $\{64,128,192,256,512\}$ and $c$ over 
$\{512,768,1024,1536\}$.
All possible combinations of $(e,q,c)$ are again evaluated 
in a grid search on the small benchmark subset (Appendix \ref{app:data}), and the best
performing model is selected for the next iteration.
We run three iterations and obtain models $\N_0$, $\N_1$, and $\N_2$.

\begin{table}[t]
   \caption{Model training and evaluation for anonymous GBDTs ($D_i$) and GNN ($\N_i$).}
\label{tab:results}
\center
\bgroup
\setlength\tabcolsep{1.5mm}
\begin{tabular}{c|ccccc|cc|cc}
   \multirow{2}{*}{$\M$} &
   TPR & 
   TNR & 
   \multicolumn{3}{c|}{training} & 
   \multicolumn{2}{c|}{real time} & \multicolumn{2}{c}{abstract time} 
\\
   &  
   [\%] & 
   [\%] & 
   size & 
   time & 
   params & 
   $\S\oplus\M$ & 
   $+\%$ & 
   $\S\oplus\M$ & 
   $+\%$ 
\\
\hline
$\emptyset$ & - & - & - &  - & - &  \num{14966} & \phantom{1}0.0 &  \num{10679} & \phantom{11}0.0 \\
\hline
$\D_0$ & 84.9 & 68.4 & 14M & 2h29m & X,d12 & \num{20679} & 38.1 & \num{17917} & \phantom{1}67.8 \\
$\D_1$ & 79.0 & 79.5 & 29M & 4h33m & X,d12 & \num{23280} & 58.2 &  \num{20760} & \phantom{1}94.4 \\
$\D_2$ & 80.5 & 79.2 & 47M & \phantom{0h}40m & L,d30,l1800 &  \num{24347} & 62.7 &  \num{22661} & 112.2 \\
\hline
$\N_0$ & 92.1 & 77.1 & 14M & \phantom{0d}17h & e20,q128,c512 &  \num{20912} & 39.7 &  \num{19755} & \phantom{1}84.9 \\
$\N_1$ & 90.0 & 78.6 & 31M & 1d19h & e10,q128,c512 &  \num{23156} & 54.7 &  \num{21737} & 103.5 \\
$\N_2$ & 91.3 & 79.6 & 50M & 1d\phantom{0}8h & e50,q256,c768 &  \num{23262} & 55.4 &  \num{22169} & 107.6 \\
\end{tabular}
\egroup
\end{table}

\vspace{1mm}
\para{Results}
are presented in Table~\ref{tab:results}.
For each model $\D_i$ and $\N_i$ we show 
(1) true positive/negative rates,
(2) training data sizes, 
(3) train times, and 
(4) the best performing parameters from the grid search.
Furthermore, for each model $\M$ we show the performance of $\S\oplus\M$ in
(5) real and (6) abstract time.
Details follow.
(1) Model accuracies are computed on samples extracted from problems newly
solved by each model, that is, on testing data not known during 
the training.
Columns TPR/TNR show accuracies on positive/negative testing samples.
(2) Train sizes measure the training data in millions of clauses.
(4) Letter ``X'' stands for XGBoost models, while ``L'' for
LightGBM.
(5) For real time we use $\SI{10}{\second}$ limit per problem, and (6) in
abstract time we limit the number of generated clauses to $\num{5000}$.
We show the number of problems solved and the gain (in
\%) on $\S$.
The abstract time evaluation is useful to assess the methods modulo the speed of
the implementation.
The first row shows the performance of $\S$ without learning.

\vspace{1mm}
\para{Evaluation.}
The GNN models start better, but the GBDT models catch up and beat GNN in later iterations.
The GBDT models show a significant gain even in the 3rd
iteration, while the GNN models start stagnating.
The GNN models report better testing accuracy, but
their ATP performance is not as good. %

For GBDTs, we see that the first two best models ($\D_0$ and $\D_1$) were
produced by XGBoost, while $\D_2$ by LightGBM.
While both libraries can provide similar results, LightGBM is
significantly faster.
For comparison, the training time for XGBoost in the third iteration
was 7 hours, that is, LightGBM is 10 times faster.
The higher speed of LightGBM can overcome the problems with more complicated
parameter settings, as more models can be trained and evaluated.

For GNNs, we observe higher training times and better models coming from
earlier epochs.
The training in the 1st and 2nd iterations was done on 1 GPU, while in the 3rd on 4 GPUs.
The good abstract time performance indicates that further gain could
be obtained by a faster implementation. But note that this is the
first time that NNs have been made comparable to GBDTs in real
time.

\begin{figure}[t]
\begin{center}
\hspace*{-0.2cm}
\includegraphics[scale=0.6]{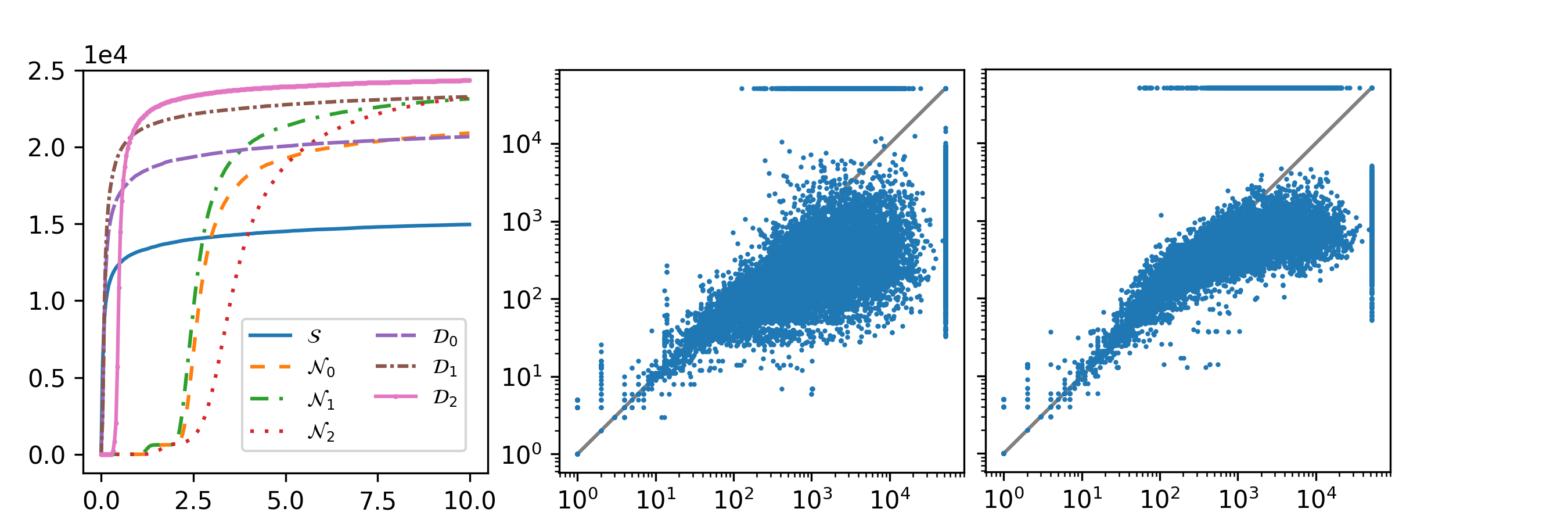}
\end{center}
\caption{Left: the number of problems solved in time;
   Right: the number of processed clauses (the $x$-axis for $\S$,
   and the $y$-axis for $\S\oplus\D_0$ and $\S\oplus\N_0$, respectively).%
}
\label{fig:cactus-scatter-nudle}
\end{figure}

Figure~\ref{fig:cactus-scatter-nudle} summarizes the results.
On the left, we observe a slower start for GNNs caused by the initial model loading.
On the right, we see a decrease in the number of processed clauses, which suggests
that the guidance is effective.

\vspace{1mm} \para{Complementarity.}  The twelve (solo and
cooperative) versions of the methods compared in
Figure~\ref{fig:cactus-scatter-nudle} solve together 28271 problems,
with the six GBDTs solving 25255 and the six GNNs solving 26571. All
twenty methods tested by us solve 29118 problems, with the top-6
greedy cover solving (in 60 s) 28067 and the top-15 greedy cover
solving (in 150 s) 29039.  The GNNs show higher complementarity --
varying the epoch as well as the size of the query and context
produces many new solutions. For example, the most complementary GNN method adds to the best GNN method 1976 solutions.
The GNNs are also quite complementary to
the GBDTs. The second (GNN) strategy in the greedy cover
adds 2045 solutions to the best (GBDT) strategy. Altogether, the
twenty strategies solve (in 200 s) 2109 of the Mizar40 \emph{hard}
problems, i.e., the problems unsolved by any method developed
previously in~\cite{KaliszykU13b}.

\section{Conclusion}

We have developed and evaluated symbol-independent GBDT and GNN
ATP guidance.  This is the first time symbol-independent features
and GNNs are tightly integrated with E and provide good real-time results on a large corpus.
Both the GBDT and GNN predictors display high ability to learn from previous proof
searches even in the symbol-independent setting.

To provide competitive real-time performance of the GNNs, we have developed
context-based evaluation of generated clauses in E.
This introduces a new paradigm for clause ranking and selection in saturation-style proving.
The generated clauses are not ranked immediately and independently of other clauses.
Instead, they are judged in larger batches and with respect to a large number of already
selected clauses (context) by a neural network that estimates their collectively most useful subset
by several rounds of message passing. This also allows new ways of parameterizing the search that result in
complementary methods with many new solutions.

The new GBDTs show even better performance than their symbol-dependent versions from
our previous work~\cite{JakubuvU19}.
This is most likely because of the parameter grid search and new features not
used before.
The union of the problems solved by the twelve ENIGMA strategies (both
$\odot$ and $\oplus$) in real time adds up to \num{28247}.
When we add $\S$ to this portfolio we solve \num{28271} problems. 
This shows that the ENIGMA strategies learned quite well from $\S$, not losing
many solutions. When we add eight more strategies developed here we solve \num{29130} problems, of which \num{2109} are among the hard Mizar40.
This is done in general in 200 s and without any additional help from premise selection methods.
Vampire in 300 seconds solves \num{27842} problems.
Future work includes joint evaluation of the system on problems translated
from different ITP libraries, similar to~\cite{DBLP:conf/lpar/GauthierK15}.

\section{Acknowledgments}

We thank Stephan Schulz and Thibault Gauthier for discussing with us
their methods for symbol-independent term and formula matching.

\bibliographystyle{plain}

\bibliography{ate11}

\begin{thebibliography}{10}

\bibitem{tensorflow2015-whitepaper}
Mart\'{\i}n Abadi, Ashish Agarwal, Paul Barham, Eugene Brevdo, Zhifeng Chen,
  Craig Citro, Greg~S. Corrado, Andy Davis, Jeffrey Dean, Matthieu Devin,
  Sanjay Ghemawat, Ian Goodfellow, Andrew Harp, Geoffrey Irving, Michael Isard,
  Yangqing Jia, Rafal Jozefowicz, Lukasz Kaiser, Manjunath Kudlur, Josh
  Levenberg, Dandelion Man\'{e}, Rajat Monga, Sherry Moore, Derek Murray, Chris
  Olah, Mike Schuster, Jonathon Shlens, Benoit Steiner, Ilya Sutskever, Kunal
  Talwar, Paul Tucker, Vincent Vanhoucke, Vijay Vasudevan, Fernanda Vi\'{e}gas,
  Oriol Vinyals, Pete Warden, Martin Wattenberg, Martin Wicke, Yuan Yu, and
  Xiaoqiang Zheng.
\newblock {TensorFlow}: Large-scale machine learning on heterogeneous systems,
  2015.
\newblock Software available from tensorflow.org.

\bibitem{BlanchetteGKKU16}
Jasmin~Christian Blanchette, David Greenaway, Cezary Kaliszyk, Daniel
  K{\"{u}}hlwein, and Josef Urban.
\newblock A learning-based fact selector for {Isabelle/HOL}.
\newblock {\em J. Autom. Reasoning}, 57(3):219--244, 2016.

\bibitem{hammers4qed}
Jasmin~Christian Blanchette, Cezary Kaliszyk, Lawrence~C. Paulson, and Josef
  Urban.
\newblock Hammering towards {QED}.
\newblock {\em J. Formalized Reasoning}, 9(1):101--148, 2016.

\bibitem{Chen:2016:XST:2939672.2939785}
Tianqi Chen and Carlos Guestrin.
\newblock {XGBoost}: A scalable tree boosting system.
\newblock In {\em Proceedings of the 22nd ACM SIGKDD International Conference
  on Knowledge Discovery and Data Mining}, KDD '16, pages 785--794, New York,
  NY, USA, 2016. ACM.

\bibitem{ChvalovskyJ0U19}
Karel Chvalovsk{\'{y}}, Jan Jakubuv, Martin Suda, and Josef Urban.
\newblock {ENIGMA-NG:} efficient neural and gradient-boosted inference guidance
  for {E}.
\newblock In Pascal Fontaine, editor, {\em Automated Deduction - {CADE} 27 -
  27th International Conference on Automated Deduction, Natal, Brazil, August
  27-30, 2019, Proceedings}, volume 11716 of {\em Lecture Notes in Computer
  Science}, pages 197--215. Springer, 2019.

\bibitem{DBLP:journals/jar/CzajkaK18}
Lukasz Czajka and Cezary Kaliszyk.
\newblock Hammer for {Coq}: Automation for dependent type theory.
\newblock {\em J. Autom. Reasoning}, 61(1-4):423--453, 2018.

\bibitem{Fan:2008:LLL:1390681.1442794}
Rong-En Fan, Kai-Wei Chang, Cho-Jui Hsieh, Xiang-Rui Wang, and Chih-Jen Lin.
\newblock Liblinear: A library for large linear classification.
\newblock {\em J. Mach. Learn. Res.}, 9:1871--1874, June 2008.

\bibitem{DBLP:conf/cpp/GauthierK15}
Thibault Gauthier and Cezary Kaliszyk.
\newblock Premise selection and external provers for {HOL4}.
\newblock In Xavier Leroy and Alwen Tiu, editors, {\em Proceedings of the 2015
  Conference on Certified Programs and Proofs, {CPP} 2015, Mumbai, India,
  January 15-17, 2015}, pages 49--57. {ACM}, 2015.

\bibitem{DBLP:conf/lpar/GauthierK15}
Thibault Gauthier and Cezary Kaliszyk.
\newblock Sharing {HOL4} and {HOL} light proof knowledge.
\newblock In Martin Davis, Ansgar Fehnker, Annabelle McIver, and Andrei
  Voronkov, editors, {\em Logic for Programming, Artificial Intelligence, and
  Reasoning - 20th International Conference, {LPAR-20} 2015, Suva, Fiji,
  November 24-28, 2015, Proceedings}, volume 9450 of {\em Lecture Notes in
  Computer Science}, pages 372--386. Springer, 2015.

\bibitem{DBLP:journals/jsc/GauthierK19}
Thibault Gauthier and Cezary Kaliszyk.
\newblock Aligning concepts across proof assistant libraries.
\newblock {\em J. Symb. Comput.}, 90:89--123, 2019.

\bibitem{GauthierKU16}
Thibault Gauthier, Cezary Kaliszyk, and Josef Urban.
\newblock Initial experiments with statistical conjecturing over large formal
  corpora.
\newblock In Andrea Kohlhase, Paul Libbrecht, Bruce~R. Miller, Adam Naumowicz,
  Walther Neuper, Pedro Quaresma, Frank~Wm. Tompa, and Martin Suda, editors,
  {\em Joint Proceedings of the FM4M, MathUI, and ThEdu Workshops, Doctoral
  Program, and Work in Progress at the Conference on Intelligent Computer
  Mathematics 2016 co-located with the 9th Conference on Intelligent Computer
  Mathematics {(CICM} 2016), Bialystok, Poland, July 25-29, 2016}, volume 1785
  of {\em {CEUR} Workshop Proceedings}, pages 219--228. CEUR-WS.org, 2016.

\bibitem{10.1007/978-3-030-29026-9_21}
Zarathustra Goertzel, Jan Jakub{\r{u}}v, and Josef Urban.
\newblock {ENIGMAWatch: ProofWatch} meets {ENIGMA}.
\newblock In Serenella Cerrito and Andrei Popescu, editors, {\em Automated
  Reasoning with Analytic Tableaux and Related Methods}, pages 374--388, Cham,
  2019. Springer International Publishing.

\bibitem{JakubuvU17a}
Jan Jakubuv and Josef Urban.
\newblock {ENIGMA:} efficient learning-based inference guiding machine.
\newblock In Herman Geuvers, Matthew England, Osman Hasan, Florian Rabe, and
  Olaf Teschke, editors, {\em Intelligent Computer Mathematics - 10th
  International Conference, {CICM} 2017, Edinburgh, UK, July 17-21, 2017,
  Proceedings}, volume 10383 of {\em Lecture Notes in Computer Science}, pages
  292--302. Springer, 2017.

\bibitem{JakubuvU18}
Jan Jakubuv and Josef Urban.
\newblock Enhancing {ENIGMA} given clause guidance.
\newblock In Florian Rabe, William~M. Farmer, Grant~O. Passmore, and Abdou
  Youssef, editors, {\em Intelligent Computer Mathematics - 11th International
  Conference, {CICM} 2018, Hagenberg, Austria, August 13-17, 2018,
  Proceedings}, volume 11006 of {\em Lecture Notes in Computer Science}, pages
  118--124. Springer, 2018.

\bibitem{JakubuvU19}
Jan Jakubuv and Josef Urban.
\newblock Hammering {Mizar} by learning clause guidance.
\newblock In John Harrison, John O'Leary, and Andrew Tolmach, editors, {\em
  10th International Conference on Interactive Theorem Proving, {ITP} 2019,
  September 9-12, 2019, Portland, OR, {USA}}, volume 141 of {\em LIPIcs}, pages
  34:1--34:8. Schloss Dagstuhl - Leibniz-Zentrum f{\"{u}}r Informatik, 2019.

\bibitem{holyhammer}
Cezary Kaliszyk and Josef Urban.
\newblock Learning-assisted automated reasoning with {F}lyspeck.
\newblock {\em J. Autom. Reasoning}, 53(2):173--213, 2014.

\bibitem{ckju-mcs-hh}
Cezary Kaliszyk and Josef Urban.
\newblock {HOL(y)Hammer}: Online {ATP} service for {HOL Light}.
\newblock {\em Mathematics in Computer Science}, 9(1):5--22, 2015.

\bibitem{KaliszykU13b}
Cezary Kaliszyk and Josef Urban.
\newblock {MizAR 40 for Mizar 40}.
\newblock {\em J. Autom. Reasoning}, 55(3):245--256, 2015.

\bibitem{LightGBM}
Guolin Ke, Qi~Meng, Thomas Finley, Taifeng Wang, Wei Chen, Weidong Ma, Qiwei
  Ye, and Tie{-}Yan Liu.
\newblock Lightgbm: {A} highly efficient gradient boosting decision tree.
\newblock In {\em {NIPS}}, pages 3146--3154, 2017.

\bibitem{Vampire}
Laura Kov{\'a}cs and Andrei Voronkov.
\newblock First-order theorem proving and {V}ampire.
\newblock In Natasha Sharygina and Helmut Veith, editors, {\em CAV}, volume
  8044 of {\em LNCS}, pages 1--35. Springer, 2013.

\bibitem{DBLP:journals/corr/abs-1911-12073}
Miroslav Ols{\'{a}}k, Cezary Kaliszyk, and Josef Urban.
\newblock Property invariant embedding for automated reasoning.
\newblock {\em CoRR}, abs/1911.12073, 2019.

\bibitem{Overbeek:1974:NCA:321812.321814}
Ross~A. Overbeek.
\newblock A new class of automated theorem-proving algorithms.
\newblock {\em J. ACM}, 21(2):191--200, April 1974.

\bibitem{DBLP:books/daglib/0002958}
Stephan Schulz.
\newblock {\em Learning search control knowledge for equational deduction},
  volume 230 of {\em {DISKI}}.
\newblock Infix Akademische Verlagsgesellschaft, 2000.

\bibitem{DBLP:conf/ki/Schulz01}
Stephan Schulz.
\newblock Learning search control knowledge for equational theorem proving.
\newblock In Franz Baader, Gerhard Brewka, and Thomas Eiter, editors, {\em {KI}
  2001: Advances in Artificial Intelligence, Joint German/Austrian Conference
  on AI, Vienna, Austria, September 19-21, 2001, Proceedings}, volume 2174 of
  {\em Lecture Notes in Computer Science}, pages 320--334. Springer, 2001.

\bibitem{Sch02-AICOMM}
Stephan Schulz.
\newblock {E - A Brainiac Theorem Prover}.
\newblock {\em AI Commun.}, 15(2-3):111--126, 2002.

\bibitem{Schulz:IJCAR-2012}
Stephan Schulz.
\newblock {Fingerprint Indexing for Paramodulation and Rewriting}.
\newblock In Bernhard Gramlich, Ulrike Sattler, and Dale Miller, editors, {\em
  Proc.\ of the 6st IJCAR, Manchester}, volume 7364 of {\em LNAI}, pages
  477--483. Springer, 2012.

\bibitem{DBLP:conf/birthday/Schulz13}
Stephan Schulz.
\newblock Simple and efficient clause subsumption with feature vector indexing.
\newblock In {\em Automated Reasoning and Mathematics}, volume 7788 of {\em
  Lecture Notes in Computer Science}, pages 45--67. Springer, 2013.

\bibitem{Urban06}
Josef Urban.
\newblock {MPTP} 0.2: Design, implementation, and initial experiments.
\newblock {\em J. Autom. Reasoning}, 37(1-2):21--43, 2006.

\bibitem{US+08-long}
Josef Urban, Geoff Sutcliffe, Petr Pudl{\'a}k, and Ji\v{r}\'{\i} Vysko\v{c}il.
\newblock {MaLARea SG1 - Machine Learner for Automated Reasoning with Semantic
  Guidance}.
\newblock In Alessandro Armando, Peter Baumgartner, and Gilles Dowek, editors,
  {\em IJCAR}, volume 5195 of {\em LNCS}, pages 441--456. Springer, 2008.

\bibitem{Veroff96}
Robert Veroff.
\newblock Using hints to increase the effectiveness of an automated reasoning
  program: Case studies.
\newblock {\em J. Autom. Reasoning}, 16(3):223--239, 1996.

\end{thebibliography}

\appendix
\section{Additional Data From the Experiments}
\label{app:data}
\begin{figure}[p]
\begin{center}
\hspace*{-1cm}
\includegraphics[angle=90,origin=c,height=0.8\paperwidth]{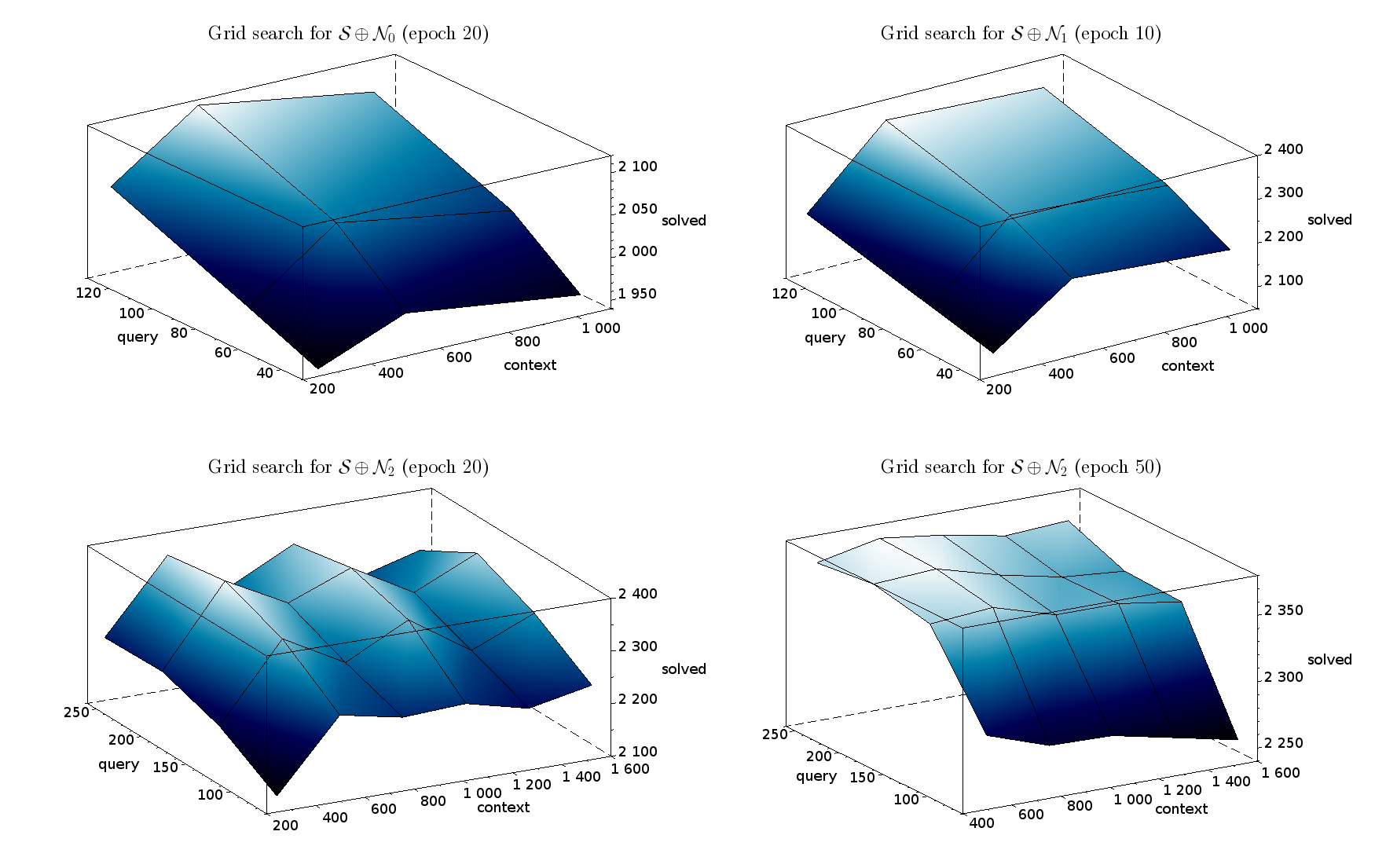}
\end{center}
\caption{Grid search results for GNN models ($\N_i$).}
\label{fig:gnn-grid}
\end{figure}

This appendix presents additional data from the experiments in
Section~\ref{sec:exp}.
Figure~\ref{fig:gnn-grid} shows the results of the grid search for GNN models on one tenth of
all benchmark problems done in order to find the best-performing parameters for
\emph{query} and \emph{context} sizes.
The $x$-axis plots the query size, the $y$-axis plots the context size, while the
$z$-axis plots the ATP performance, that is, the number of solved problems.
Recall that the grid search was performed on a randomly selected but fixed
tenth of all benchmark problems with a \SI{10}{\second} real-time limit per
problem.
For $\N_0$ and $\N_1$, there is a separate graph for each iteration, showing
only the best epochs.
For $\N_2$, there are two graphs for models from epoch 20 and 50.
Note how the later epoch 50 becomes more independent on the context size.
The ranges of the grid search parameters were extended in later iterations when
the best-performing value was at the graph edge.

\begin{figure}[p]
\begin{center}
\hspace*{-1cm}
\includegraphics[angle=90,origin=c,height=0.8\paperwidth]{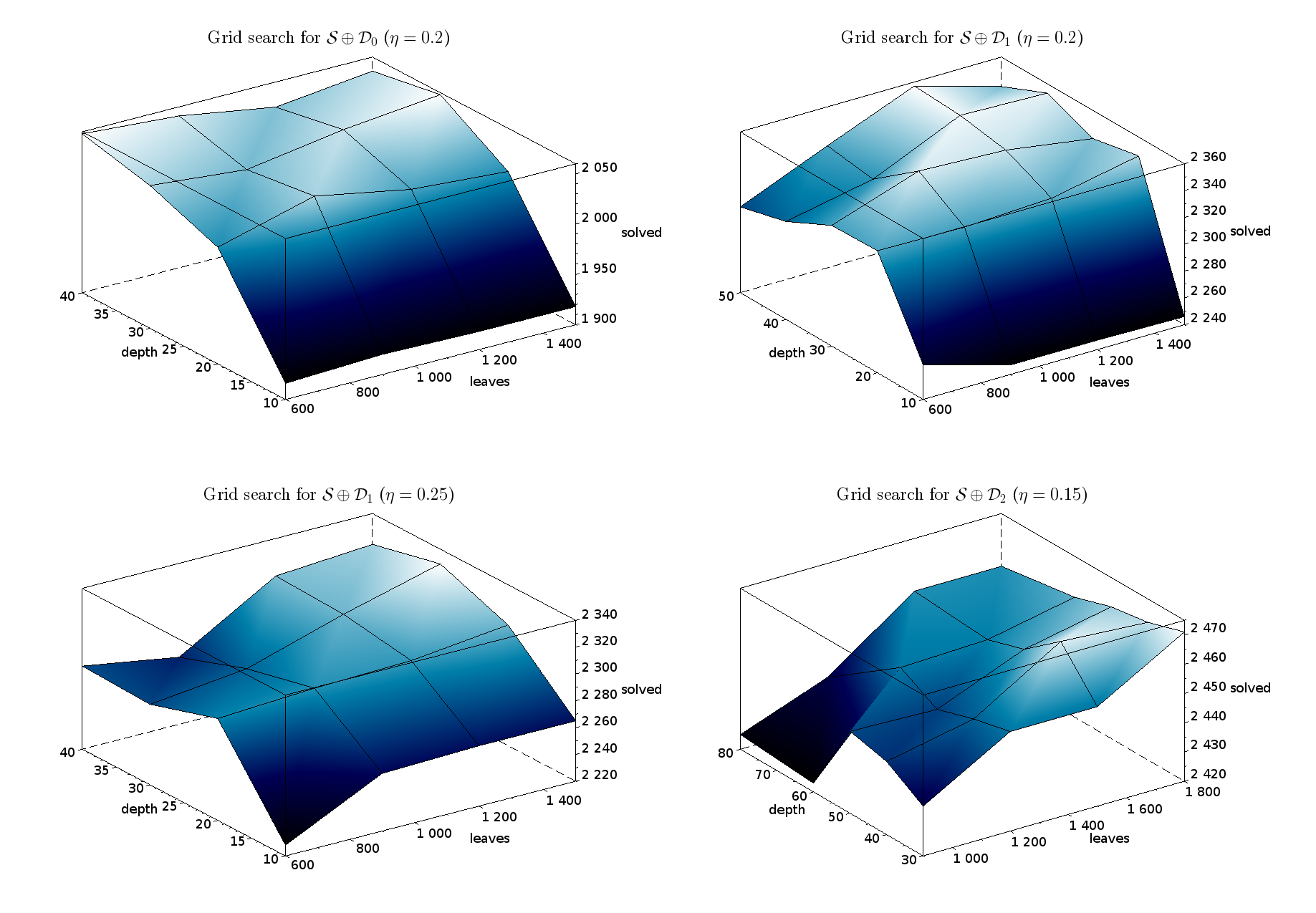}
\end{center}
\caption{Grid search results for LightGBM GBDT models ($\D_i$).}
\label{fig:lgb-grid}
\end{figure}

Figure~\ref{fig:lgb-grid} shows the grid search results for the best LightGBM's
GBDT models from iterations $1$, $2$, and $3$ (denoted here $\D_0$, $\D_1$, and
$\D_2$).
The $x$-axis plots the number of tree leaves, the $y$-axis plots the tree depth,
while the $z$-axis plots the number of solved problems.
There are two models from the second iteration ($\D_1$), showing the effect of
different learning rate ($\eta$).
Again, the ranges of meta-parameters were updated in between the
iterations by a human engineer.

\begin{figure}[p]
\begin{center}
\hspace*{-1cm}
\includegraphics[angle=90,origin=c,height=0.85\paperwidth]{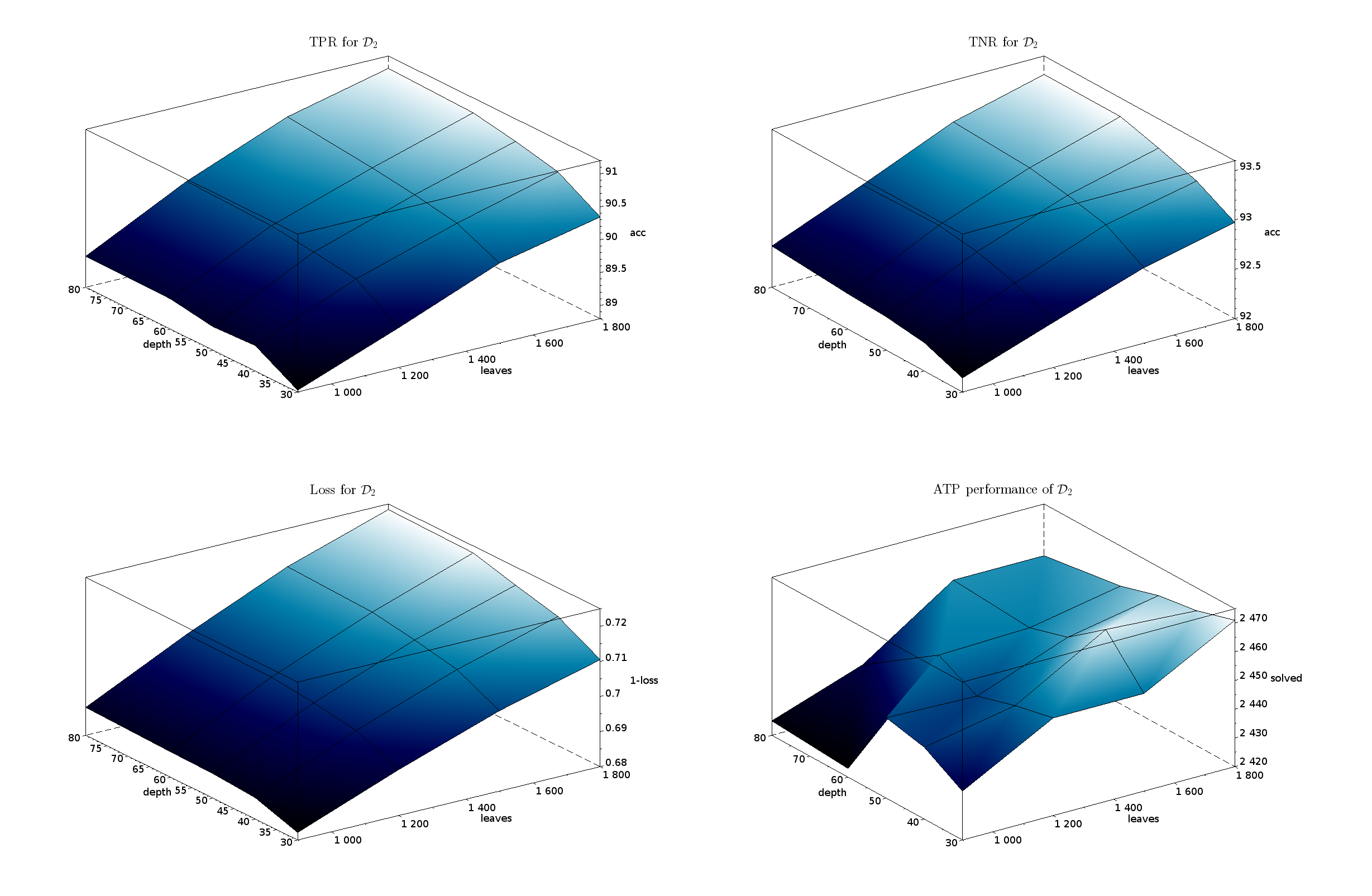}
\end{center}
\caption{Accuracies for LightGBM model $\D_2$.}
\label{fig:lgb-acc}
\end{figure}

Figure~\ref{fig:lgb-acc} shows the training accuracies and training loss for
the LightGBM model $\D_2$.
Accuracies (TPR and TNR) of the training data are computed from the first
iteration ($\T_0$).
The values for loss ($z$) are inverted ($1-z$) so that higher values correspond
to better models which makes a visual comparison easier. 
We can see a clear correlation between the accuracies and the loss, but not so
clear correlation with the ATP performance.
The ATP performance of $\D_2$ is the same as in Figure~\ref{fig:lgb-grid},
repeated here for convenience.

\begin{figure}[h]
\begin{center}
\hspace*{-1cm}
\includegraphics[scale=0.6]{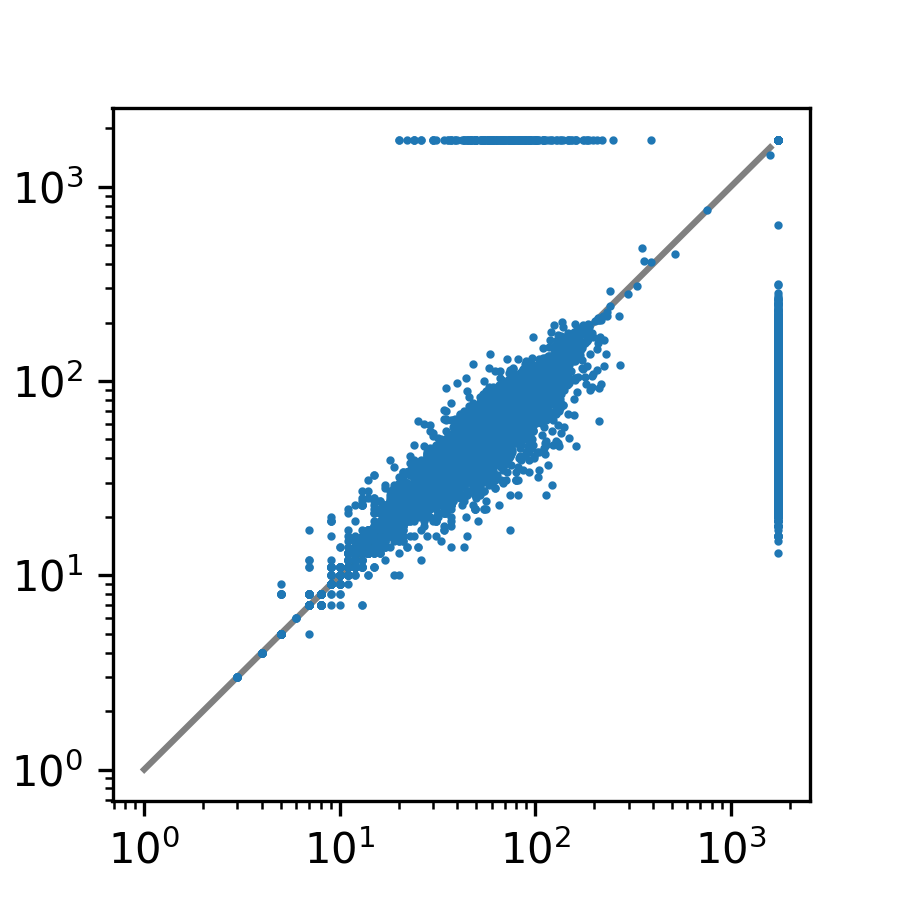}
\includegraphics[scale=0.6]{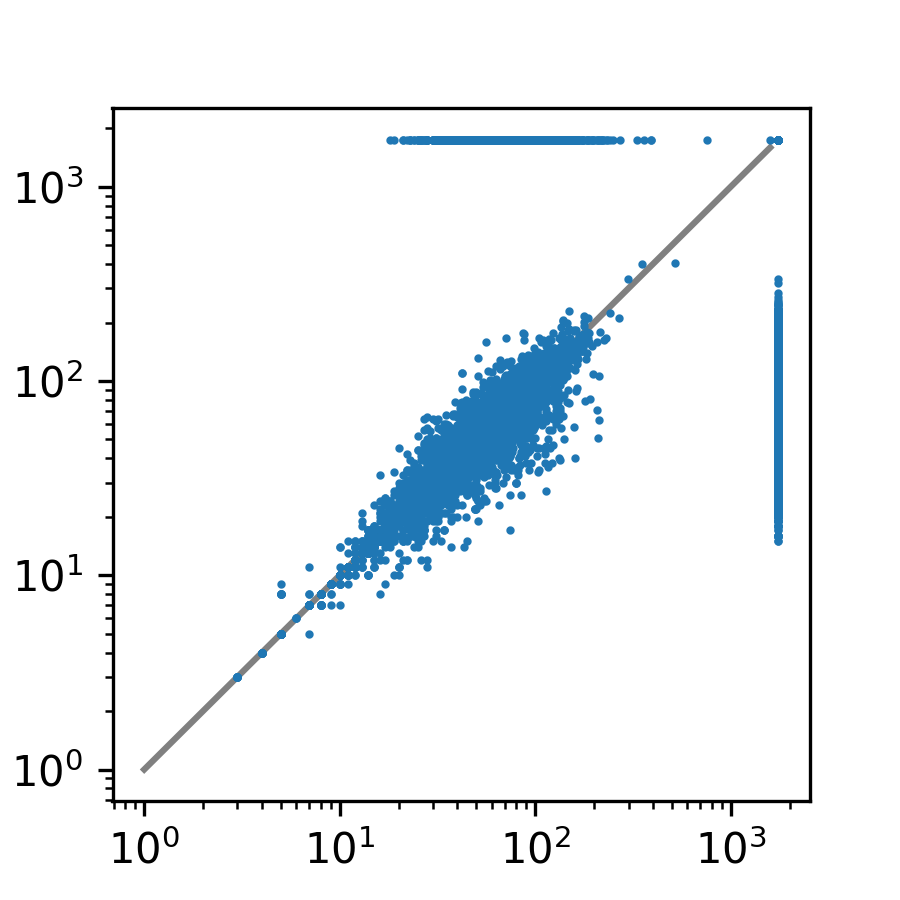}
\end{center}
\caption{Scatter plots for the lengths of the discovered proofs
	 (the $x$-axis for $\S$, and the $y$-axis for $\S\oplus\D_2$ and $\S\oplus\N_2$, respectively).}
\label{fig:prooflenght}
\end{figure}

Figure~\ref{fig:prooflenght} compares the lengths of the discovered proofs.
We can see that there is no systematic difference in this metric between the base strategy 
and the ENIGMA ones.

Finally, we have compared the feature vectors of the symbol-dependent and
symbol-independent versions of the GBDTs. On the same data, we observe roughly 2x more
collisions. %
The symbol-independent version has around 1\% of colliding feature vectors, while the symbol-dependent version has 0.42\%. 

\section{Discussion of Anonymization}
\label{app:anon}
Our use of symbol-independent arity-based features for GBDTs
differs from Schulz's anonymous \emph{clause patterns}~\cite{DBLP:conf/ki/Schulz01,DBLP:books/daglib/0002958} (CPs)
used in E for proof
guidance and from Gauthier and Kaliszyk's (GK) anonymous abstractions used for
their concept alignments between ITP libraries~\cite{DBLP:journals/jsc/GauthierK19} in two ways:
\begin{enumerate}
\item  In both CP and GK, serial (de Bruijn-style) numbering of
abstracted symbols of the same arity is used. I.e., the term $h(g(a))$
will get abstracted to $F11(F12(F01))$. Our encoding is just
$F1(F1(F0))$. It is even more lossy, because it is the same for $h(h(a))$.

\item  ENIGMA with gradient boosting decision trees (GBDTs) can be
(approximately) thought of as implementing weighted feature-based
clause classification where the feature weights are learned. Whereas
both in CP and GK, exact matching is used after the abstraction is
done.\footnote{We thank Stephan Schulz for pointing out that although CPs used exact matching by default, matching up to a certain depth was also implemented.}
In CP, this is used for hint-style guidance of E. There, for
clauses, such serial numbering however isn't stable under literal
reordering and subsumption. Partial heuristics can be used, such as 
normalization based on a fixed global ordering done in both CP and GK.
\end{enumerate}

Addressing the latter issue (stability under reordering of literals and
subsumption) leads to the NP hardness of (hint)
matching/subsumption. I.e., the abstracted subsumption task can be
encoded as standard first-order subsumption for clauses where terms
like $F11(F12(F01))$ are encoded as \\ $apply1(X1,apply1(X2,apply0(X3)))$.  The NP hardness of subsumption is however
here more serious in practice than in standard ATP because only
applications behave as non-variable symbols during the matching.

Thus, the difference between our anonymous approach and CP is
practically the same as between the standard symbol-based ENIGMA
guidance and standard hint-based~\cite{Veroff96} guidance. In the former the matching
(actually, clause classification) is approximate, weighted and
learned, while with hints the clause matching/classification is crisp,
logic-rooted and preprogrammed, sometimes running into the NP hardness
issues. Our latest comparison~\cite{10.1007/978-3-030-29026-9_21} done over the Mizar/MPTP corpus in the symbol-based setting
showed better performance of ENIGMA over using hints, most likely due to better generalization behavior of ENIGMA based on the statistical (GBDT) learning.

Note also that the variable and symbol statistics features to some extent
alleviate the conflicts obtained with our encoding. E.g., $h(g(a))$ and
$h(h(a))$ will have different \emph{symbol statistics} (Section~\ref{sec:gbdt}) features. To some
extent, such features are similar to Schulz's feature vector and fingerprint indexing~\cite{DBLP:conf/birthday/Schulz13,Schulz:IJCAR-2012}.

\end{document}